# Onboard Multivariable Controller Design for a Small Scale Helicopter Using Coefficient Diagram Method

Agus Budiyono

*Department of Aeronautics and Astronautics*

*Bandung Institute of Technology*

*Jl. Ganesha 10, Bandung 40132, Indonesia*

Electronic mail: agus.budiyono@ae.itb.ac.id

**Abstract.** A mini scale helicopter exhibits not only increased sensitivity to control inputs and disturbances, but also higher bandwidth of its dynamics. These properties make model helicopters, as a flying robot, more difficult to control. The dynamics model accuracy will determine the performance of the designed controller. It is attractive in this regards to have a controller that can accommodate the unmodeled dynamics or parameter changes and perform well in such situations. Coefficient Diagram Method (CDM) is chosen as the candidate to synthesize such a controller due to its simplicity and convenience in demonstrating integrated performance measures including equivalent time constant, stability indices and robustness. In this study, CDM is implemented for a design of multivariable controller for a small scale helicopter during hover and cruise flight. In the synthesis of MIMO CDM, good design common sense based on hands-on experience is necessary. The low level controller algorithm is designed as part of hybrid supervisory control architecture to be implemented on an onboard computer system. Its feasibility and performance are evaluated based on its robustness, desired time domain system responses and compliance to hard-real time requirements.

## Introduction

The control for a small scale helicopter has been designed using various methods. During the period of 1990s, the classical control systems such as single-input-single-output SISO proportional-derivative (PD) feedback control systems have been used extensively. Their controller parameters were usually tuned empirically. This trial-and-error approach to design an "acceptable" control system however is not agreeable with complex multi-input multi-output MIMO systems with sophisticated performance criteria. For more advanced multivariable controller synthesis approaches, an accurate model of the dynamics is required. Such models, however, are not usually readily available and are difficult to develop [1]. To control a model helicopter as a complex MIMO system, an approach that can synthesize a control algorithm to make the helicopter meet performance criteria while satisfying some physical constraints is required. More recent development in this area include the use of optimal control (Linear Quadratic Regulator) implemented on a small aerobatic helicopter designed at MIT[2-3]. Similar approach based on $\mu$ -synthesis has been also independently developed for a rotor unmanned aerial vehicle at UC Berkeley[4]. An adaptive high-bandwidth helicopter controller algorithm was synthesized at Georgia Tech.[5].

Based on the analysis of the simulation results as well as experimental verifications[4], it has been shown that the MIMO approach is superior to the SISO design due to the fact that the coupling between variables is inherently handled. The most widely used MIMO approaches, such as LQR and $H_\infty$, however have drawbacks which make them not very amenable to practical implementation in general. These include higher than necessary order of controller, non-existence of formal parameter tuning and weight selection procedures, possible exclusion of good controllers, and difficulty in integrating state variable constraints[6]. CDM has been developed fairly recently to address this need. The technique can be viewed as *generalized* PID where its basic principles have been known for over than 40 years in various segments of industries including servo control, aircraft automatic control, gas turbine control, robotics manipulators, spacecraft attitude control, DC motor control and many others. CDM is an algebraic approach applied to polynomial loop in the parameter space, where the so-called coefficient diagram is used as the means to convey the necessary design information and as the criteria of good design.

Thus far CDM approach has been successfully applied primarily to single variable control synthesis. A very limited attempt has been made to extend its potential application to multivariable system. To the best of the author's knowledge, beyond the pioneering work by Manabe in [6-7], there has been only very limited recorded results for the attempt in this direction such as the CDM application for steel mill drive control [8]. The current paper addresses this challenge and reports some progress towards CDM application to multivariable control synthesis for high bandwidth dynamics of a small scale helicopter.

## Dynamics of a Small Scale Helicopter

The Yamaha R-50 helicopter dynamics model has been developed at Carnegie Mellon Robotics Institute. It uses a two-bladed main rotor with a Bell-Hiller stabilizer bar. The physical characteristic of the helicopter is summarized in the following Table 1[1]:

Table 1

Physical parameter of R-50 Helicopter

| | |
|---|---|
| Rotor Speed | 850 rpm |
| Tip speed | 449 ft/s |
| Dry weight | 97 lb |
| Instrumented | 150 lb |
| Engine Single cylinder | 2-stroke |
| Flight autonomy | 30 minutes |

The basic linearized equations of motion for a model helicopter dynamics are derived from the Newton-Euler equations for a rigid body that has six degrees of freedom to move in space. The external forces, consisting aerodynamic and gravitational forces, are represented in a stability derivative form. For simplicity, the control forces produced by the main and tail rotor are expressed by the multiplication of a control derivative and the associated control input. Following [1], the equations of motion of the model helicopter are derived and categorized into the following groups.

**Fuselage dynamics.** Using the Newton-Euler equations, the translational and angular fuselage motions of the helicopter can be derived as the following set of equations:

$$
\begin{aligned}
\dot{u} &= (-w_0 q + v_0 r) - g\theta + X_u u + \cdots + X_a a \\
\dot{v} &= (-u_0 r + w_0 p) - g\phi + Y_v v + \cdots + Y_b b \\
\dot{w} &= (-v_0 p + u_0 q) + Z_w w + Z_{col}\delta_{col} \\
\dot{p} &= L_u u + L_v v) + \cdots + L_b b \\
\dot{q} &= M_u u + M_v v) + \cdots + M_a a \\
\dot{r} &= N_r r + N_{ped}(\delta_{ped} - r_{fb}) \\
\dot{r}_{fb} &= K_r r - K_{rfb} r_{fb}
\end{aligned}
\tag{1}
$$

**Coupled Rotor-stabilizer dynamics.** The simplified rotor dynamics is represented by two first-order differential equations for lateral (b) and longitudinal (a) flapping motion. The state-space model is given as follows:

$$
\begin{aligned}
\tau_f \dot{b} &= -b - \tau_f p + B_a a + B_{lat}\delta_{lat} + B_d d + B_{lon}\delta_{lon} \\
\tau_f \dot{a} &= -a - \tau_f q + A_b b + A_{lat}\delta_{lat} + A_c c + A_{lon}\delta_{lon}
\end{aligned}
\tag{2}
$$

The fly-bar dynamics similarly is represented by similar equations for lateral (d) and longitudinal (c) flapping motion as follows:

$$
\begin{aligned}
\tau_s \dot{d} &= -d - \tau_s p + D_{lat}\delta_{lat} \\
\tau_s \dot{c} &= -c - \tau_s q + C_{lon}\delta_{lon}
\end{aligned}
\tag{3}
$$

**The state-space model of R-50 helicopter.** The state space model of the helicopter can be assembled from the above set of differential equations in a matrix form:

$$
\dot{\underline{x}} = A\underline{x} + B\underline{u} \tag{4}
$$

where $\underline{x} = \{u, v, w, \theta, \phi, p, q, r, a, b, r_{fb}, c, d\}$ is the state vector and $\underline{u} = \{\delta_{lat}, \delta_{lon}, \delta_{ped}, \delta_{col}\}$ the input vector. The dynamic matrix $A$ contains the stability derivatives and the control matrix $B$ contains the input derivatives. The complete description the elements of these matrices are presented in [1].

## CDM Procedures

**Mathematical model.** The mathematical model of the CDM design is described in general as a block diagram shown in Fig.[1]. In this figure, $r$ is the reference input signal, $u$ is the control signal, $d$ is the disturbance and $n$ is the noise generated by the measuring device at the output; N(s) and D(s) are the numerator and denominator polynomial of the plant transfer function, respectively. A(s), F(s) and B(s) are the polynomials associated with the CDM controller which are the denominator polynomial matrix of the controller, the reference and the feedback numerator polynomial matrix of the controller respectively. For MIMO case, the variables and components are in the form of vectors and matrices with the appropriate dimension.

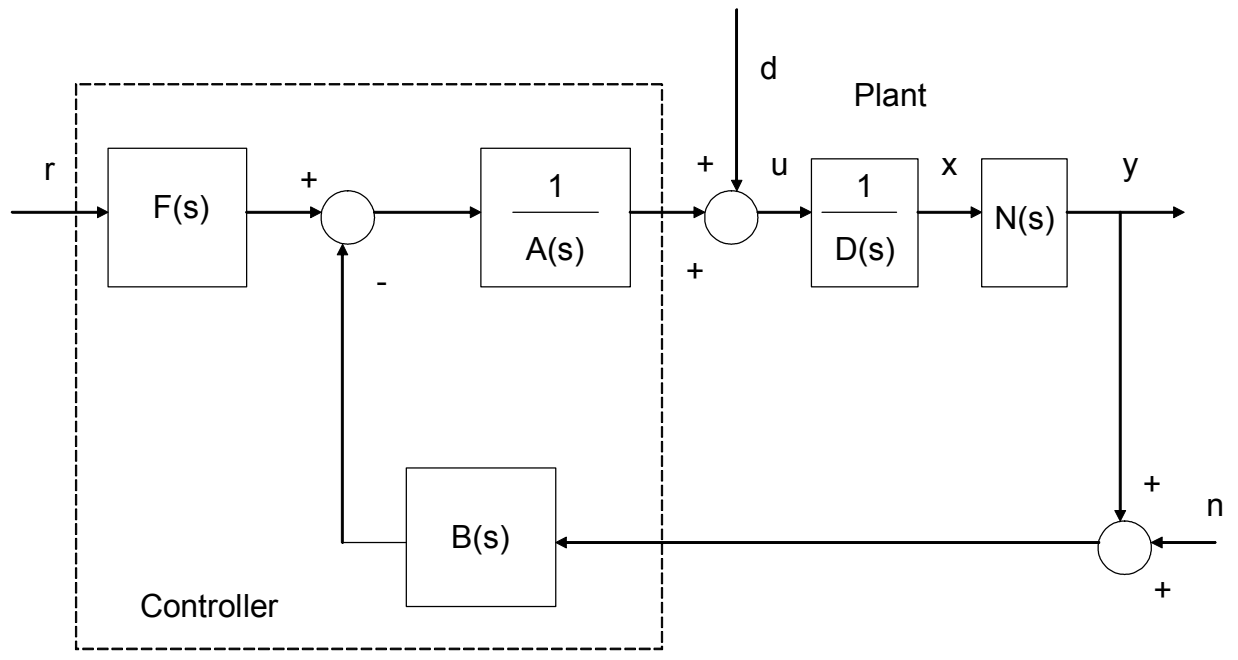

Fig. 1 CDM block diagram

The plant equation is given by:

$$y = N(s)x$$

$$y = \frac{N(s)}{D(s)}(u + d) \tag{5}$$

which after some algebraic manipulation, can be completely written as:

(6)

$$y = \frac{\mathrm{N(s)F(s)}}{\mathrm{P(s)}} r + \frac{\mathrm{A(s)N(s)}}{\mathrm{P(s)}} d - \frac{\mathrm{N(s)B(s)}}{\mathrm{P(s)}} n \tag{7}$$

where P(s) is the closed-loop system polynomial matrix expressed by:

$$P(s) = A(s)D(s) + B(s)N(s) = \sum_{i=0}^{n} a_i s^i \tag{8}$$

The characteristic polynomial $\Delta(s)$ is given by:

$$\Delta(s) = \det P(s) \tag{9}$$

To write the input-output relation of the system, the expression for the state and the controllers are needed. The controller equation can be written as:

$$A(s)u = F(s)r - B(s)(n + y) \tag{10}$$

Whereas the state equation can be obtained by eliminating u and y from the controller and output equations as follows:

$$P(s)x = F(s)r + A(s)d - B(s)n \tag{11}$$

Combining the output, state and controller equations, Eqs.(5),(11) and (10), the matrix input-output equation can finally be expressed as:

$$\begin{bmatrix} x \\ y \\ z \end{bmatrix} = \frac{1}{\Delta(s)} \begin{bmatrix} I \\ N(s) \\ D(s) \end{bmatrix} adjA(s)[F(s)r + A(s)d - B(s)n] - \begin{bmatrix} 0 \\ 0 \\ d \end{bmatrix} \tag{12}$$

**CDM controller design.** The design parameters in CDM are the stability indices $\gamma_i 's$, the stability limit indices $\gamma_i^* 's$ and the equivalent time constant, $\tau$. The stability index and the stability limit index determine the system stability and the transient behavior of the time domain response. In addition, they determine the robustness of the system to parameter variations. The equivalent time constant, which is closely related to the bandwidth, determines the rapidity of the time response. Those parameters are defined as follows:

$$\begin{aligned} &\gamma_i = \frac{a_i^2}{(a_{i+1}a_{i-1})}, \quad i = 1, 2, \cdots, n-1 \\ &\tau = \frac{a_1}{a_0} \\ &\gamma_i^* = \frac{1}{\gamma_{i+1}} + \frac{1}{\gamma_{i-1}}, \quad \gamma_0 = \gamma_n \triangleq \infty \end{aligned} \tag{13}$$

where $a_i$'s are coefficients of the characteristic polynomial $\Delta(s)$. The equivalent time constant of the i-th order $\tau_i$ is defined in the same way as $\tau$.

$$\tau_i = \frac{a_{i+1}}{a_i} \tag{14}$$

By using the above equations, the relation between $\tau_i$'s can be written as:

$$\frac{\tau_i}{\tau_{i-1}} = \frac{a_{i+1}}{a_i}\frac{a_{i-1}}{a_i} = \frac{1}{\gamma_i} \tag{15}$$

Also, by simple manipulation, $a_i$ can be written as:

$$\begin{aligned} &a_i = \tau_{i-1} \cdots \tau_1 \tau a_0 \\ &a_i = \frac{a_0 \tau^i}{\gamma_{i-1}\gamma^2_{i-2} \cdots \gamma^{i-2}_2 \gamma^{i-1}_1}, \quad i \ge 2 \end{aligned} \tag{16}$$

The characteristic polynomial can then be expressed as:

$$\Delta(s) = a_0 \left[ \left\{ \sum_{i-2}^{n} \left( \prod_{j=1}^{i-1} \frac{1}{\gamma^j_{i-j}} \right) (\tau s)^i \right\} + \tau s + 1 \right] \tag{17}$$

The sufficient condition for stability is given as:

$$\begin{aligned} &a_i > 1.12 \left[ \frac{a_{i-1}}{a_{i+1}} a_{i+2} + \frac{a_{i+1}}{a_{i-1}} a_{i-2} \right] \\ &\gamma_i > 1.12 \gamma_i^*, \forall i = 2, 3, \cdots, n-2 \end{aligned} \tag{18}$$

And the sufficient condition for instability is:

$$
\begin{aligned}
&a_{i+1}a_i \le a_{i+2}a_{i-1} \\
&\gamma_{i+1}\gamma_i \le 1, \quad \text{for some } i = 1, \cdots, n-2
\end{aligned}
\tag{19}
$$

## Feedback Controller Design for Model Helicopter

A common approach in the modeling of helicopter dynamics is separating the equations of motion into two parts: the longitudinal-vertical and lateral-directional modes. In this work, CDM has been implemented for both modes. However, due to space limitation, the reported results are restricted for the controller design in the longitudinal-vertical mode only. The state-space model of the longitudinal-vertical mode is given by:

$$
\begin{bmatrix} \dot{u} \\ \dot{q} \\ \dot{\theta} \\ \dot{a} \\ \dot{w} \end{bmatrix} = \begin{bmatrix} X_u & 0 & -g & X_a & 0 \\ M_u & 0 & 0 & M_a & M_w \\ 0 & 1 & 0 & 0 & 0 \\ 0 & -1 & 0 & -\frac{1}{\tau_f} & 0 \\ 0 & 0 & 0 & Z_a & Z_w \end{bmatrix} \begin{bmatrix} u \\ q \\ \theta \\ a \\ w \end{bmatrix} + \begin{bmatrix} 0 & 0 \\ 0 & M_{col} \\ 0 & 0 \\ \frac{A_{lon}}{\tau_f} & 0 \\ 0 & Z_{col} \end{bmatrix} \begin{bmatrix} \delta_{lon} \\ \delta_{col} \end{bmatrix}
\tag{20}
$$

where the numerical values of the parameters and stability derivatives, both for hover and cruise flight, are given in [1]. The outputs of the system are assumed to be all the state variables except the flapping motion $a$.

The goal of the control design is essentially to maintain helicopter at hovering position i.e. the well-known hover control problem. In this design, CDM will be applied to first ensure the stability of the closed-loop system. The open-loop system is known to be unstable as shown later. Specifically, the helicopter should follow a commanded forward velocity reference, $u_{ref}$, using the available control inputs.

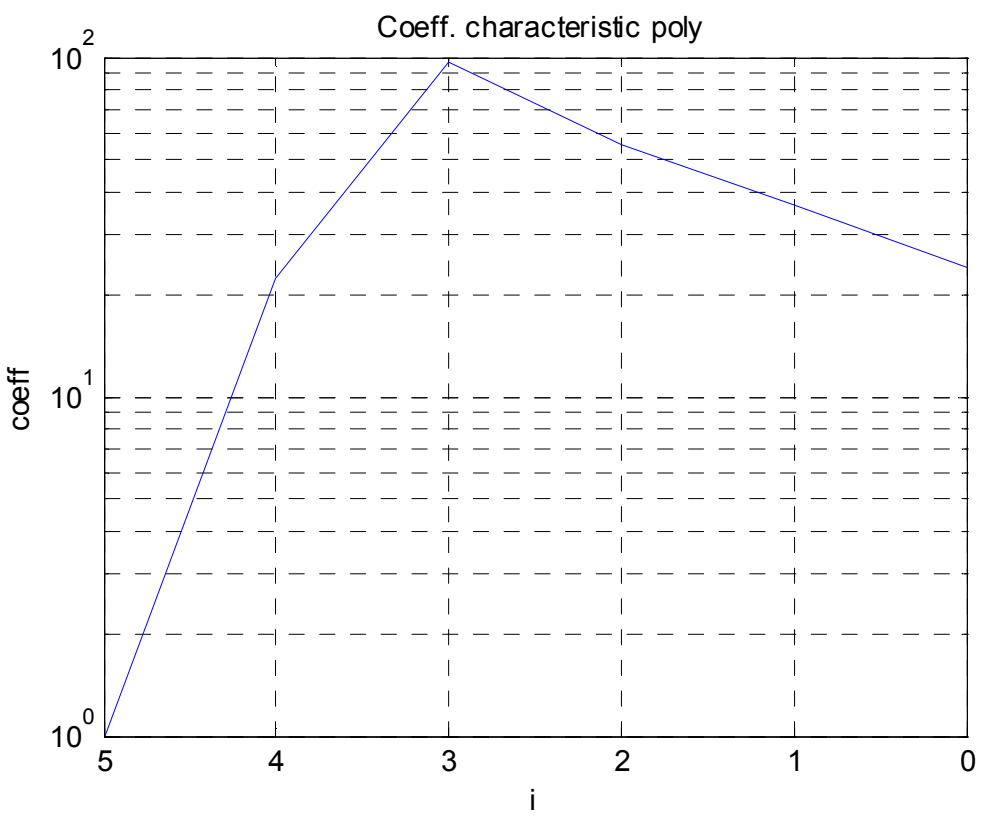

Fig. 2 Coefficient Diagram of C-L polynomial

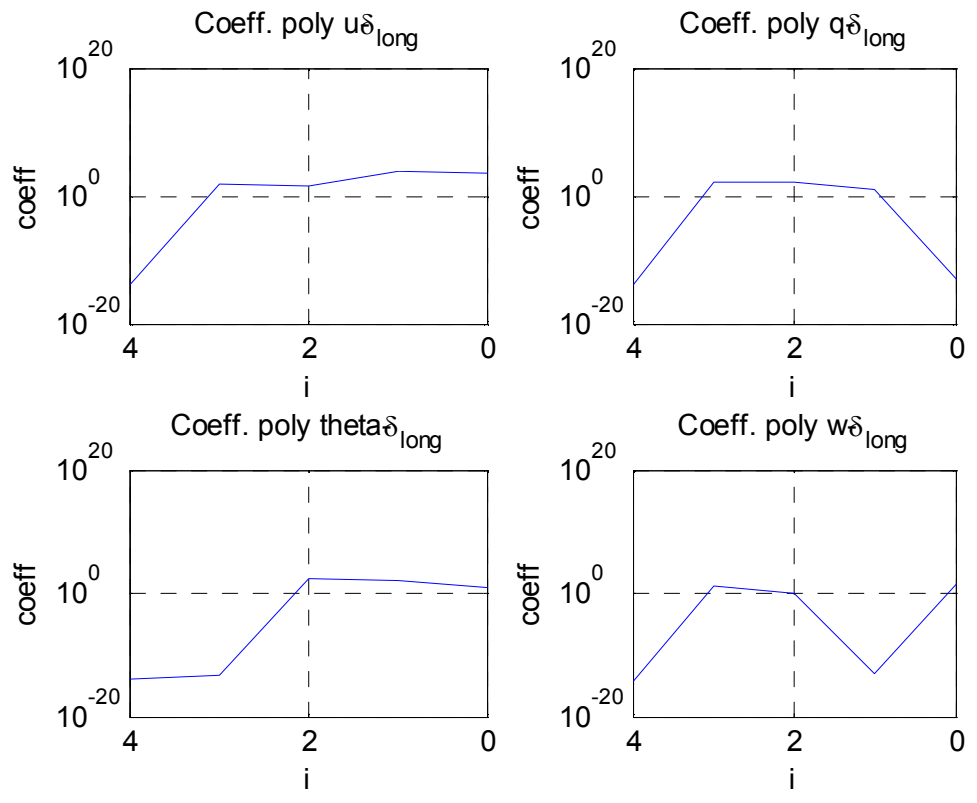

Fig. 3 Coefficient diagram for input-output relation

Using the standard CDM procedures, the characteristic polynomial and eight input-output relations are derived as the following:

$$
\begin{aligned}
&\Delta(s) = s^5 + 22.4s^4 + 97.08s^3 + 55.56s^2 - 36.71s - 24.11 \\
&(u \to \delta_{lon})(s) = \varepsilon s^4 + 70s^3 + 43s^2 + 5782s + 3550.1 \\
&(q \to \delta_{lon})(s) = -179.56s^3 - 123.25s^2 - 7.98s - \varepsilon \\
&(\theta \to \delta_{lon})(s) = \varepsilon s^4 - \varepsilon s^3 + 179.56s^2 - 123.25s - 7.98 \\
&(w \to \delta_{lon})(s) = \varepsilon s^4 + 21.2s^3 + 1.07s^2 + \varepsilon s - 38.29 \\
&(w \to \delta_{col})(s) = -45.8s^4 - 998s^3 - 3833.4s^2 - 191s + 1798.6 \\
&(u \to \delta_{col}) = (q \to \delta_{col}) = (\theta \to \delta_{col}) = 0
\end{aligned} \tag{21}
$$

The coefficient diagram of the characteristic polynomial and $\delta_{lon}$-output relation are shown in Fig. 2 and 3 respectively. In Eq.(21), $\varepsilon$ is a small number that needs to be neglected to achieve usable results during parameter tuning of the controller. Using the inequality requirement for stability in Eq.(18), it can be shown that for $n = 2, a_2 = 55.56 < 61.588$ which means that the system is unstable. The control design objective in this case is to change the coefficient of polynomial in order to stabilize the system using the appropriate feedback. By observation, it is more effective to use $\delta_{lon}$ to achieve the objective since $\delta_{col}$ is effective only when the vertical velocity feedback is used. Various PID controllers were compared and evaluated. One of the designs using $u, \theta$ and $w$ feedback is shown as Simulink block diagram implementation in Fig. 4.

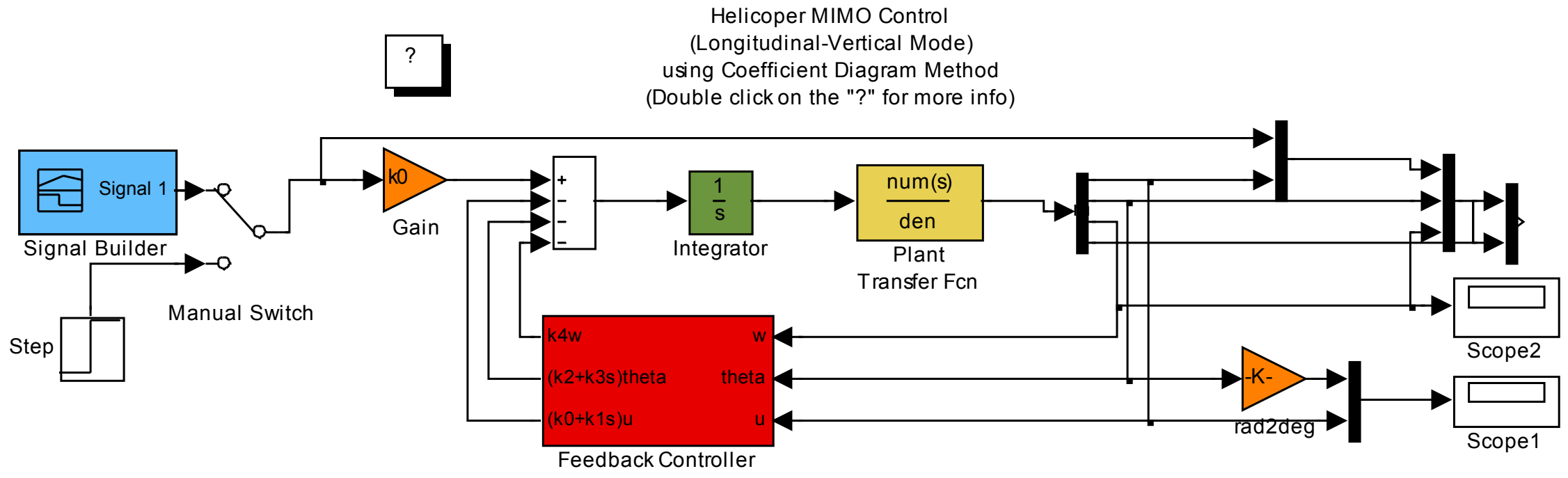

Fig.4 The feedback control design using CDM

Using the above diagram, it can be observed that the PID controller is chosen such that:

$$
s\delta_{lon} = k_0 u_r - [(k_0 + k_1 s)u + (k_2 + k_3 s)\theta + k4w] \tag{22}
$$

The new characteristic polynomial P(s) then becomes:

$$
P(s) = s\Delta(s) + (k_0 + k_1 s)(u \to \delta_{lon}) + (k_2 + k_3 s)(\theta \to \delta_{lon}) + k4(w \to \delta_{lon}) \tag{23}
$$

$$
P(s) = s^6 + a_5 s^5 + a_4 s^4 + a_3 s^3 + a_2 s^2 + a_1 s + a_0 \tag{24}
$$

Solving the Diophantine equation, the controller parameters for the feedback design shown in Fig. 4 are obtained as: k0 = 0.08412,k1 =-0.30369,k2 =-13.90378,k3 =-2.56712,k4 =2.46190 . The corresponding controller performance is shown in Fig. 5-6 together with other feedback controllers.

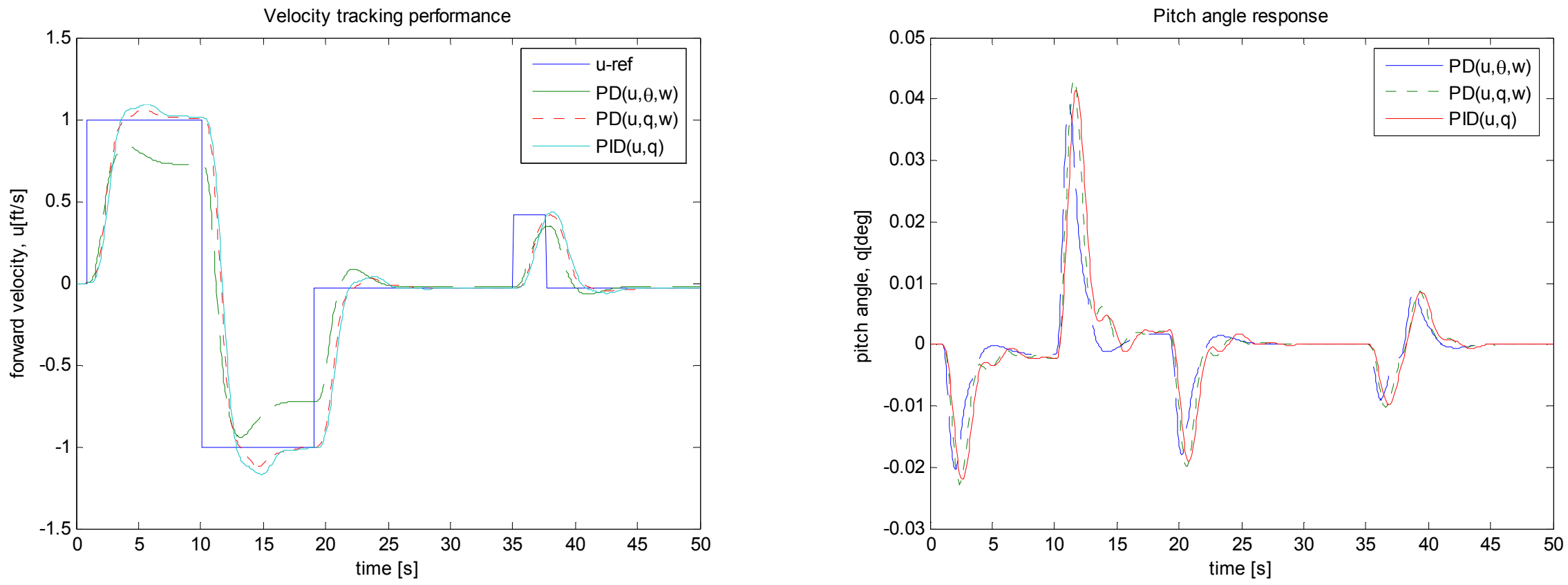

Fig. 5 The unit doublet response for forward velocity, u

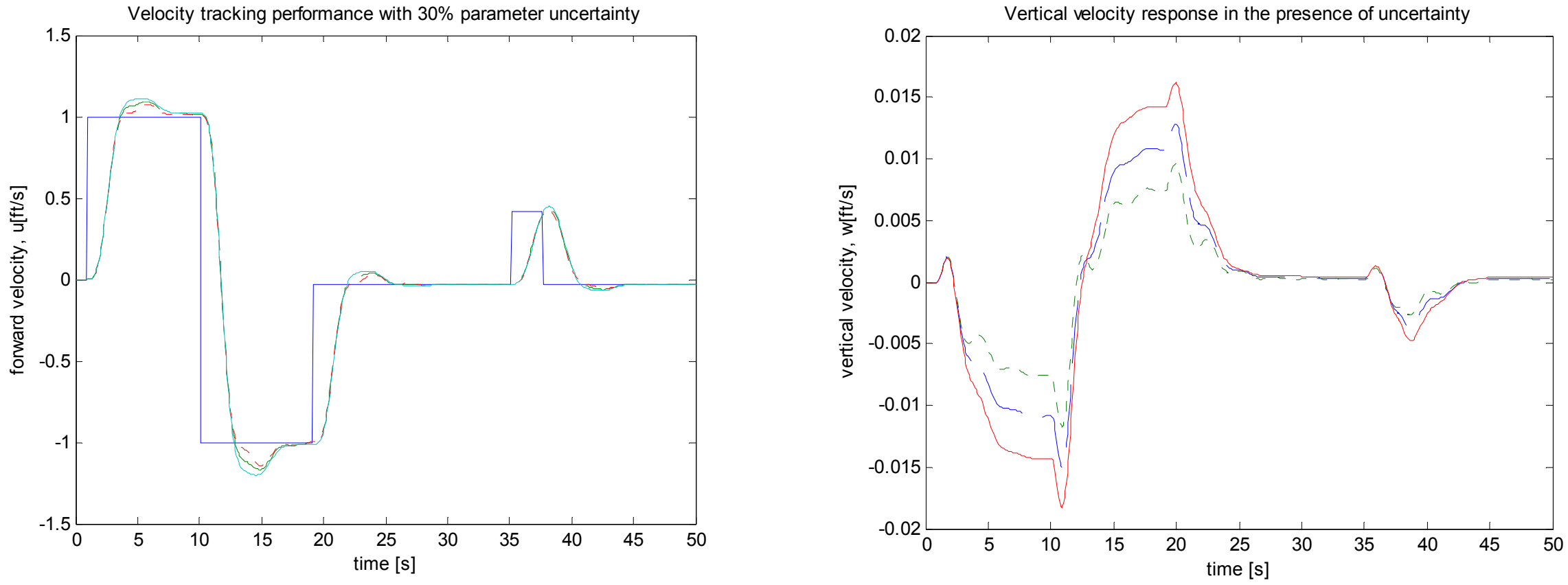

Fig. 6 The unit doublet response in the case of $\pm 30\%$ uncertainty in the $X_u, X_a, M_u$ and $M_w$

## Discussion on Results

The responses of a unit doublet in the input and to an impulse disturbance at t=35s are given in Fig.5. It is evident that feedbacks involving pitch rate have better performance than that with pitch angle. The controllers design using CDM all show a good disturbance rejection with zero steady-state error. The output is completely controlled by the reference input. Fig. 6 demonstrates the effects of $\pm 30\%$ uncertainty in the helicopter stability parameters $X_u, X_a, M_u$ and $M_w$ on doublet response with the same disturbance at t=35s. The controller design using CDM with forward velocity and pitch rate feedback is used in this investigation. It can be observed that the controller is fairly robust under both disturbance and additional parameter perturbations. The corresponding settling time is practically unaffected. The results show the viability of the practical hover control of a model helicopter using CDM.

## Concluding Remarks

The control for a small scale helicopter has been designed using CDM which is an algebraic approach to control synthesis. It has been demonstrated that the CDM controller is robust under the effects of disturbances as wells parameter variations. It will be attractive to further investigate the optimality of the controller in comparison with the optimal control design as in [9].The low level controller algorithm is designed as part of hybrid supervisory control architecture to be implemented on an onboard computer system of an autonomous model helicopter currently under development at Bandung Institute of Technology. The controlled system is simulated and tested within the Matlab/Simulink environment. The simplicity of the approach and the designed controller promise a good compliance to existing hard real-time requirements in the future real plant implementation. The designed algorithm will be implemented in the Hardware-In-the-Loop environment in the near future.

## Acknowledgement

The work was partially supported by the A2 Program Grant of the Directorate General of Higher Education of Indonesia. The support from Dr. Bambang K. Hadi as the Head of Aeronautics and Astronautics Department for the ongoing research program is gratefully acknowledged. The author would like to thank H.Y. Sutarto for the discussions regarding the approach to robust control design.